\documentclass[11pt,a4paper]{article}

\usepackage{times,latexsym}
\usepackage{url}

\usepackage{latexsym}
\usepackage{booktabs}
\usepackage{soul,color}
\usepackage{todonotes} 
\usepackage[normalem]{ulem}

\usepackage[T1]{fontenc}

\usepackage{enumitem}

    \usepackage[acceptedWithA]{tacl2018v2}
\usepackage[]{tacl2018v2}

\usepackage{amsfonts,amsmath,amsthm}
\theoremstyle{plain}

\usepackage{mathtools}

\defcitealias{joulin2015inferring}{J\&M (2015)}

\newcommand{\weight}[1] {\mathbf{W}_{#1}}
\newcommand{\bias}[1] {\mathbf{b}_{#1}}
\newcommand{\hidden}[1] {\mathbf{h}_{#1}}
\newcommand{\hiddenp}[1] {\mathbf{\tilde{h}}_{#1}}

\usepackage{xspace,mfirstuc,tabulary}

\newif\iftaclinstructions
\taclinstructionsfalse 
\iftaclinstructions
\renewcommand{\confidential}{}
\renewcommand{\anonsubtext}{(No author info supplied here, for consistency with
TACL-submission anonymization requirements)}
\newcommand{\instr}
\fi

\iftaclpubformat 

\else

\fi

\title{Memory-Augmented Recurrent Neural Networks \\ Can Learn Generalized Dyck Languages}

\author{Mirac Suzgun\textsuperscript{1} \hfill Sebastian Gehrmann\textsuperscript{1} \hfill Yonatan Belinkov\textsuperscript{12} \hfill Stuart M. Shieber\textsuperscript{1} \\\\ 
  \textsuperscript{1} Harvard John A. Paulson School of Engineering and Applied Sciences \\  
  \textsuperscript{2} MIT Computer Science and Artificial Intelligence Laboratory \\ 
  Cambridge, MA, USA \\ 
  \texttt{\{msuzgun@college,\{gehrmann,belinkov,shieber\}@seas\}.harvard.edu} }

\date{}
\newcommand{\dyck}[1]{\ensuremath{\mathcal{D}_{#1}}}
\newcommand{\gendyck}{\dyck{>1}}

\begin{document}

\maketitle

\begin{abstract}
We introduce three memory-augmented Recurrent Neural Networks (MARNNs) and explore their capabilities on a series of simple language modeling tasks whose solutions require stack-based mechanisms. We provide the first demonstration of neural networks recognizing the generalized Dyck languages, which express the core of what it means to be a language with hierarchical structure. Our memory-augmented architectures are easy to train in an end-to-end fashion and can learn the Dyck languages over as many as six parenthesis-pairs, in addition to two deterministic palindrome languages and the string-reversal transduction task, by emulating pushdown automata. Our experiments highlight the increased modeling capacity of memory-augmented models over simple RNNs, while inflecting our understanding of the limitations of these models. 
\end{abstract}

\section{Introduction}
Recurrent Neural Networks (RNNs) have proven to be an effective and powerful model choice for capturing long-distance dependencies and complex representations in sequential tasks, such as language modeling \citep{mikolov2010recurrent,Sundermeyer2012LSTMNN}, machine translation \citep{kalchbrenner2013recurrent, sutskever2014sequence,bahdanau2014neural}, and speech recognition \citep{graves2013speech}. In theory, RNNs with rational state weights and infinite numeric precision are known to be computationally universal models~\citep{siegelmann1994analog, siegelmann1995computational}. Yet, in practice, the computational power of RNNs with finite numeric precision is still unknown. Hence, the classes of languages that can be learned, empirically or theoretically, by RNNs with finite numeric precision are still to be discovered. 

A natural question arises, then, as to what extent RNN models can learn languages that exemplify important formal properties found in natural languages, such properties as long-distance dependencies, counting, hierarchy, and repetition. 

Along these lines, \citet{gers2001lstm, weiss2018practical, suzgun2019evaluating} have demonstrated that Long Short-Term Memory (LSTM; \citet{hochreiter1997long}), a popular variant of RNNs, can develop \emph{counting} mechanisms to recognize simple strictly context-free and context-sensitive languages, such as $a^n b^n$ and $a^n b^n c^n$, as evidenced by analysis of the hidden state values.\footnote{From an automata-theoretic perspective, such languages are expressible with simple one-turn counter machines.} By contrast, \citeauthor{weiss2018practical} have shown that Gated Recurrent Units (GRUs; \citet{cho2014learning}), another popular variant of RNNs, cannot perform this type of counting and provided an explanation for some of the difference in performance between LSTMs and GRUs. 

\citet{merrill2019sequential} studied the theoretical expressiveness of various real-time neural networks with finite precision under asymptotic conditions, showing that RNNs and GRUs can capture regular languages whereas LSTMs can further recognize a subset of real-time counter languages. And empirically, \citet{suzgun-etal-2019-lstm} demonstrated that LSTM networks can learn to perform dynamic counting, as exemplified by the well-balanced parenthesis (Dyck) language \dyck{1} as well as the shuffles of multiple \dyck{1} languages.

{Counting} in real-time is an important property, differentiating some of the language classes in the Chomsky hierarchy, and echoes of it appear in natural languages as well, for instance, in the requirement that the number of arguments of a set of verbs match their subcategorization requirements. But counting does not exhaust the kinds of structural properties that may be apposite for natural language. \citet{Chomsky1957} emphasizes the \emph{hierarchical structure} found in natural languages, for instance, in the nested matching of ``both \ldots and'' and ``either \ldots or''. Indeed, this kind of nested-matching phenomenon forms the essence of the strictly context-free languages (CFLs). Here simple (real-time) counters are not sufficient; a stack is required. The formal-language-theory reflex of this phenomenon is found most sparely in the Dyck languages \dyck{n} of well-nested strings over $n$ pairs of brackets, where $n > 1$. (We refer to these as the \gendyck\ languages.)

The centrality of this nested stack structure in characterizing the class of context-free languages can be seen in various ways. (i) The automata-theoretic analog of context-free grammars, the pushdown automaton, is defined by its use of a stack \citep{chomsky1962context}. (ii) \citet{chomsky1963algebraic} famously showed that all context-free languages are homomorphic images of regular-intersected \dyck{n} languages.%
\footnote{In particular, \dyck{2} is sufficient for this purpose \citep{magniez2014recognizing, suzgun-etal-2019-lstm}.}
(iii) The hardest CFL of \citet{greibach1973hardest} and the hardest deterministic CFL of \citet{sudborough1976deterministic} are built using Dyck-language-style matching.
For these reasons, we think of the \gendyck\ languages as expressing the core of what it means to be a context-free language with hierarchical structure, even if it is not itself a universal CFL.
This property of the Dyck languages  accounts for the heavy focus on the them in prior work \citep{deleu2016learning, bernardy2018can,sennhauser2018evaluating, skachkova2018closing, hao2018context, zaremba2016learning, suzgun2019evaluating, yu-etal-2019-learning, hahn2019theoretical} as well as in this work. It would thus be notable for finite precision neural networks to learn languages, like the \gendyck\ languages and other languages requiring a stack, if we want these neural architectures to be able to manifest hierarchical structures.

In this paper, we introduce three enhanced RNN models that consist of recurrent layers and external memory structures, namely stack-augmented RNNs (Stack-RNNs), stack-augmented LSTMs (Stack-LSTMs), and Baby Neural Turing Machines (Baby-NTMs), and show that they can effectively learn to recognize some \gendyck\ languages from limited data by emulating deterministic pushdown automata. Previous studies used simple RNN models \citep{bernardy2018can,sennhauser2018evaluating,suzgun-etal-2019-lstm, yu-etal-2019-learning} and memory-augmented architectures \citep{hao2018context} to attempt to learn \dyck{2} under different training platforms; however, none of them were able to obtain good performance on this task. We thus present the first demonstration that a memory-augmented neural network (MARNN) can learn \gendyck\ languages.
Moreover, we evaluate the learning capabilities of our architectures on 
six tasks whose solutions require the employment of stack-based approaches, namely learning the \dyck{2}, \dyck{3} and \dyck{6} languages, recognizing the deterministic palindrome language ($w\#w^R)$ and the deterministic homomorphic palindrome language ($w\#\varphi(w^R)$), and performing the string-reversal transduction task ($w\#^{|w|} \Rightarrow \#^{|w|} w^R$). Our results reflect the better modeling capacity of our MARNNs over the standard RNN and LSTM models in capturing hierarchical representations, in addition to providing an insightful glimpse of the promise of these models for real-world natural-language processing tasks.\footnote{Our code is available at \url{https://github.com/suzgunmirac/marnns}.} 

\section{Related Work}
\vspace{-0.5em}
 \subsection{Learning Formal Languages Using Neural Networks}
Using neural network architectures to recognize formal languages has been a central computational task in gaining an understanding of their expressive ability for application to natural-language-processing tasks. \citet{elman1991distributed} marked the beginning of such methodological investigations and devised an artificial language learning platform where Simple Recurrent Networks (SRNs) \citep{elman1990finding}, were trained to learn the hierarchical and recursive relationships between clauses. An analysis of the hidden state dynamics revealed that the models learned internal representations that encoded information about the grammatical structure and dependencies of the synthetic language. Later, \citet{das1992learning} introduced the first RNN model with an external stack, the Recurrent Neural Network Pushdown Automaton (NNPDA), to learn simple deterministic context-free grammars.

Following \citeauthor{elman1991distributed}'s work, many studies used SRNs \citep{steijvers1996recurrent, tonkes1997learning,holldobler1997designing,rodriguez1998recurrent,boden1999learning, boden2000context,rodriguez2001simple} and stack-based RNNs \cite{das1993using,zeng1994discrete} to recognize simple context-free and context-sensitive counter languages, including $a^n b^n$, $a^nb^ncb^ma^m$, $a^nb^mB^mA^n$, $a^{n+m}b^nc^{m}$, $a^n b^n c^n$, $(ba^n)^m$, and \dyck{1}. Nonetheless, none of the SRN-based models were able to generalize far beyond their training set. Some of these studies also focused on understanding and visualizing the internal representations learned by the hidden units of the networks, as well as the computational capabilities and limitations of the models.

In contrast, \citet{gers2001lstm}, \citet{schmidhuber2002learning}, and \citet{gers2002learning}, showed that their (LSTM) networks could not only competently learn two strictly context-free languages, $a^n b^n$ and $a^n b^m B^m A^n$, and one strictly context-sensitive language $a^n b^n c^n$, but also generalize far beyond the training datasets. 

\vspace{-0.5em}
 \subsection{Memory-Augmented Neural Networks}
Recently, memory-augmented architectures have been considered for language modeling tasks: \citet{joulin2015inferring} proposed a differentiable stack structure controlled by an RNN to infer algorithmic patterns that require some combination of counting and memorization. Though their model could learn $a^n b^n$, $a^n b^n c^n$, $a^n b^n c^n d^n$, $a^nb^{2n}$, $a^nb^mc^{n+m}$, it did not exceed the performance of a standard LSTM on a language modeling task. Inspired by the early architecture design of NNPDA, \citet{grefenstette2015learning} introduced LSTM models equipped with unbounded differentiable memory structures, such as stacks, queues, and double-linked lists, and explored their computational power on synthetic transduction tasks. In their experiments, their Neural-Stack and Neural-Queue architectures outperformed the standard LSTM architectures. Neither of these studies on stack-augmented neural architectures inspected the internal representations learned by the recurrent hidden layers or investigated the performance of their models on the Dyck language.

\citet{graves2014neural} introduced the Neural Turing Machine (NTM), which consists of a neural network (which can be either feed-forward or recurrent) together with a differentiable external memory, and demonstrated its successful performance on a series of simple algorithmic tasks, such as copying, repeated copying, and sorting. At each time step, an NTM can interact with the external memory via its differentiable attention mechanisms and determine its output using the information from the current hidden state together with the filtered context from the external memory. It is evident from the design differences that the degree of freedom of NTMs is much greater than that of stack-augmented recurrent networks. However, this freedom comes at a price: The different ways in which we can attend to the memory to read and write at each time step make the training of the neural models challenging. Since the publication of the original NTM paper, there have been a number of studies addressing instability issues of the NTM architecture, or more broadly memory-augmented recurrent network models, and proposing new architecture designs. We refer to \citet{zaremba2015reinforcement, kurach2015neural, yang2016lie,graves2016hybrid, gulcehre2018dynamic} for such proposals.

\vspace{-0.6em}
 \subsection{Investigations of the Dyck Languages}\label{sec:related-dyck}
 \vspace{-0.2em}
\citet{deleu2016learning} used NTMs to capture long-distance dependencies in \dyck{1}. Their examination showed that NTMs indeed learn to emulate stack representations and generalize to longer sequences. However, a model need not be equipped with a stack to recognize this simplest Dyck language in a standard learning environment; counting is sufficient for an automaton to capture \dyck{1} \citep{suzgun-etal-2019-lstm, yu-etal-2019-learning}. 

In assessing the ability of recurrent neural networks to process deep and long-distance dependencies, \citet{skachkova2018closing} and \citet{sennhauser2018evaluating} conducted experiments on the Dyck languages to see whether LSTMs could learn nested structures. The former sought to predict the single correct closing parenthesis, given a Dyck word without its final closing symbol. Although LSTMs performed almost perfectly in this completion task, one cannot draw any definitive conclusion about whether these models really learn the Dyck languages, since even counter automata can achieve perfect accuracy on this task.%
\footnote{For instance, a model can learn to separately count the number of left and right parentheses for each of the $n$ distinct pairs and predict the closing parenthesis for the pair for which the counter is non-zero. \citet{suzgun-etal-2019-lstm} and \citet{yu-etal-2019-learning} also discuss the drawbacks of \citeauthor{skachkova2018closing}'s learning task and argue that the task is insufficient for illustrating that a network can learn \dyck{2}.}

Similarly, \citet{bernardy2018can} used three different recurrent networks, namely LSTM, GRU, and RUSS, and combinations thereof, to predict the next possible parenthesis at each time step, assuming that it is a closing parenthesis.
His RUSS model is a purpose-designed model containing recurrent units with stack-like states and appears to generalize well to deeper and longer sequences. However, as the author mentions, the specificity of the RUSS architecture disqualifies it as a practical model choice for real-world language modeling tasks.

\citet{hao2018context} studied the interpretability of Neural Stack models \citep{grefenstette2015learning} in a number of simple language modeling tasks, including parenthesis prediction, string reversal, and XOR evaluation. Though their Neural Stacks exhibited intuitive stack behaviors on their context-free transduction tasks and performed almost as well as the standard LSTM models, the authors noted that their stack-augmented models were more difficult to train than the traditional LSTMs.

More recently, \citet{suzgun-etal-2019-lstm} corroborated the theoretical findings of \citet{weiss2018practical} by showing that  RNN, GRU, and LSTM models could perform dynamic counting by recognizing \dyck{1} as well as shuffles of multiple \dyck{1} languages by emulating simple $k$-counter machines, while being incapable of recognizing \dyck{2}.

Adopting the experimental framework of \citet{sennhauser2018evaluating} and the data generation procedure of \citet{skachkova2018closing}, \citet{yu-etal-2019-learning} conducted experiments on \dyck{2} under different training schemes and objectives using relatively large bi-directional LSTM models. Their recurrent networks failed to generalize well beyond the scope of their training data to learn \dyck{2} under the closing-parenthesis completion and sequence-to-sequence settings.\footnote{We note that the authors attempted to generate the shortest proper sequence of closing parentheses given a prefix of a Dyck word under the sequence-to-sequence framework. This task is different from the previous tasks and requires a generative model.}

Finally, \citet{hahn2019theoretical} used \dyck{2} to explore the theoretical limitations of self-attention architectures \citep{vaswani2017attention}. He demonstrated that self-attention models, even when equipped with infinite precision, cannot capture \dyck{2}, unless the number of layers or attention heads increases with the length of the input sequence.

In summary, recognizing the Dyck languages has been an important probing task for understanding the ability of neural networks to capture hierarchical information. Thus far, none of the recurrent neural networks have been shown to capture \gendyck. This present work, therefore, provides the first demonstration of RNN-based models learning \gendyck, in particular, \dyck{2}, \dyck{3}, and \dyck{6}, in addition to other difficult context-free languages.

\section{Models}
In this section, we describe the mathematical formulations of our memory-augmented RNNs.The inspiration for our stack-augmented neural architectures came from the pushdown automaton, an abstract machine capable of recognizing context-free languages. Similar stack-based neural networks, however, have also been proposed by others \cite{pollack1991induction, das1992learning, joulin2015inferring, grefenstette2015learning}. Our models differ from them in their theoretical simplicity and empirical success. Our Baby-NTM, on the other hand, can be considered as a simplification of the original NTM architecture \citep{graves2014neural}: As opposed to using soft-attention mechanisms to read and write to the external memory, we make deterministic decisions and always read content from and write to the first entry of the memory, thereby making the learning process easier while retaining universal expressivity.

\vspace{-0.2em}
\paragraph{Notation} We will assume the following notation: 
\begin{itemize}[topsep=\parskip]
\setlength\itemsep{-0.15em}
\item $x = x_1,...,x_T$: The input sequence of one-hot vectors, with the $i$-th token $x_i$.
\item $y_i$: The output associated with $x_i$.
\item $\mathbf{W}$: The learnable weights of the model.
\item $\mathbf{b}$: The learnable bias terms of the model.
\item $\mathbf{h}_i$: The $i$-th hidden state representation.
\item $D$: The dim.\ of the input and output samples.
\item $H$: The dim.\ of the hidden state of the model.
\item $M$: The dim.\ of the external stack/memory.
\end{itemize}

\subsection{Stack-RNN}
\label{sec:stack-rnn}
Before we begin describing our Stack-RNN, recall the formulation of a standard RNN:
\vspace{-0.2em}
\begin{align*} 
\hidden{t} &= \text{tanh}(\weight{ih} x_t + \bias{ih} + \weight{hh} \hidden{(t-1)} + \bias{hh}) \\
y_t &= f(\weight{y} \hidden{t})
\end{align*}
where $x_t \in \mathbb{R}^D$ is the input, $\hidden{t} \in \mathbb{R}^H$ the hidden state, $y_t \in \mathbb{R}^D$ the output at time $t$, $\weight{y} \in \mathbb{R}^{D \times H}$ the linear output layer, and $f$ a transformation.

\begin{figure*}[t]
    \centering
    \includegraphics[width=0.85\textwidth]{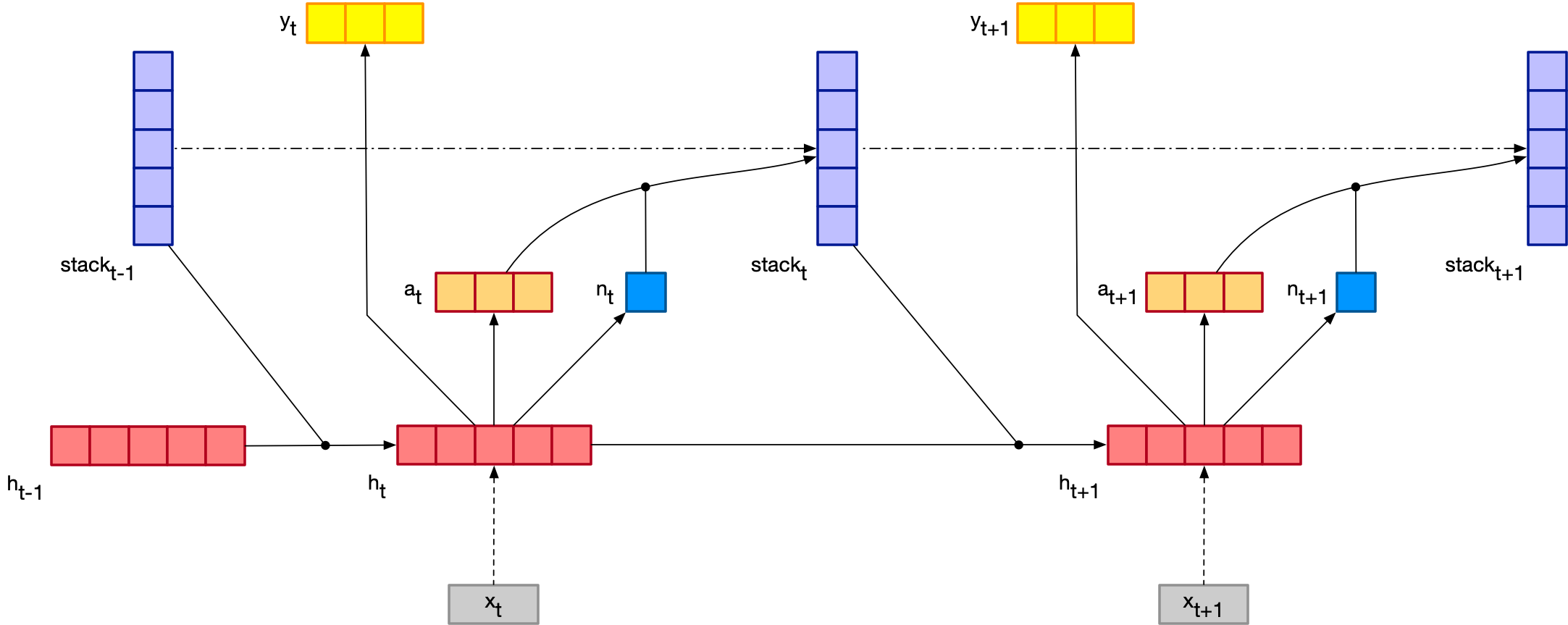}
    \caption{An abstract representation of our Stack-RNN architecture.}
    \label{fig:lstm_arch}
\end{figure*} 

While designing our Stack-RNN, we come across two important questions: (i) Where and how should we place the stack in the neural network, and (ii) how should we design the stack so that we can backpropagate errors through the stack at the time of training? Regarding (i), we place the stack in such a way that it interacts with the hidden layers at each time step. The benefit of this approach is that errors made in future stages of the model affect and backpropagate through not only the hidden states but also the stack states. Regarding (ii), we construct a \textit{differentiable} stack structure. Figure~\ref{fig:lstm_arch} provides a visualization of the Stack-RNN. Its formulation is:
\begin{align*} 
\hiddenp{(t-1)} &= \hidden{(t-1)} + \weight{sh} s_{(t-1)}^{(0)} \\
\hidden{t} &= \text{tanh}(\weight{ih} x_t + \bias{ih} + \weight{hh} \hiddenp{(t-1)} + \bias{hh})\\
y_t &= \sigma (\weight{y} \hidden{t})\\
a_t &= \text{softmax} (\weight{a} \hidden{t}) \\
n_t &= \sigma (\weight{n} \hidden{t}) \\
s_t ^{(0)} &= a_t^{(0)} n_t + a_t^{(1)} s_{(t-1)}^{(1)} \\ 
s_t^{(i)} &= a_t^{(0)} s_{(t-1)}^{(i-1)} + a_t^{(1)} s_{(t-1)}^{(i+1)}
\end{align*}
\noindent where $s_t = s_t^{(0)} s_t^{(1)} \cdots s_t^{(k)}$ is the stack configuration at time step $t$, with $s_t^{(0)}$ the topmost stack element; $\weight{sh} \in \mathbb{R}^{H \times M}$, $\weight{y} \in \mathbb{R}^{D \times H}$, $\weight{a} \in \mathbb{R}^{2 \times H}$, and $\weight{n} \in \mathbb{R}^{M \times H}$ are all learnable linear weights of the model. 

At each time step, we combine the topmost stack element $s_{(t-1)}^{(0)}$ with the previous hidden state $\hidden{(t-1)}$ via a linear mapping to produce an intermediate hidden state $\hiddenp{(t-1)}$. We then use $\hiddenp{(t-1)}$, together with the input,
to generate the current hidden state $\hidden{t}$, from which both the output at that time step $y_{t}$ and the weights of the \texttt{PUSH} ($a_{t}^{(0)}$) and \texttt{POP} ($a_{t}^{(1)}$) operations by the stack controller are determined simultaneously. Here $a_t \in \mathbb{R}^2$ is a probability distribution over the two operations. 
Finally, we update the stack elements in such a way that the elements become the weighted linear interpolation of both possible stack operations. We can, therefore, consider the elements in the stack as variables in \textit{superposition} states.

We highlight the following differences between our Stack-RNN and the Stack-RNN by \citet{joulin2015inferring}, as further explicated in the appendix. First, their model does not contain the term $\hiddenp{(t-1)}$ and it updates $\hidden{t}$ as follows:
\vspace{-0.2em}
\begin{align*}
    \hidden{t} &= \sigma(\weight{ih} x_t + \weight{hh} \hidden{(t-1)} + \weight{sh} s_{(t-1)}^{(0:k)})
\end{align*}
where $\weight{sh} \in \mathbb{R}^{H \times k}$ and $s_{(t-1)}^{(0:k)}$ the $k$-topmost elements of the stack at time $t-1$. But a simple analysis of our Stack-RNN formulation divulges that $s_{(t-1)}^{(0)}$ depends on both $\weight{hh}$ and $\weight{sh}$ in our formulation, whereas it only depends on $\weight{sh}$ in \citeauthor{joulin2015inferring}'s formulation. Furthermore, their architecture takes the sigmoid of the linear combination of $x_t$, $\hidden{(t-1)}$, and $s_{(t-1)}^{(0:k)}$, in addition to excluding the bias terms, to update $\hidden{t}$.

\subsection{Stack-LSTM}
The Stack-LSTM is similar to the Stack-RNN but contains additional components of the standard LSTM architecture by \citet{hochreiter1997long}. In this model, we update the hidden state of the model according to the standard LSTM equations, that is $\hidden{t} = \text{LSTM} (x_t, \hiddenp{(t-1)})$.

\subsection{Baby-NTM}
The Baby-NTM is both an extension of the Stack-RNN and a simplification of the original NTM. While the Stack-RNN contains an unbounded stack mechanism, it can perform only two basic operations on the stack, namely the \texttt{PUSH} and \texttt{POP} actions. In the Baby-NTM architecture, we fix the size of the external memory but provide more freedom to the model: While the interaction between the controller and the memory in the design of the Baby-NTM is mostly similar to that of the Stack-RNN, we allow five operations on the memory at each time step to update its contents: \texttt{ROTATE-RIGHT}, \texttt{ROTATE-LEFT}, \texttt{NO-OP}, \texttt{POP-LEFT}, and \texttt{POP-RIGHT}. Suppose that the current memory configuration $\mathbf{M}$ is $[a, b, c, d, e]$, where $\mathbf{M}^{(i)} \in \mathbb{R}$. Then the operations produce the following configurations at the next time step: 
\begin{align*}
\texttt{ROTATE-RIGHT}&: [e, a, b, c, d]. \\
\texttt{ROTATE-LEFT}& : [b, c, d, e, a]. \\
\texttt{NO-OP} &: [a, b, c, d, e]. \\
\texttt{POP-RIGHT} &: [0, a, b, c, d]. \\
\texttt{POP-LEFT} &: [b, c, d, e, 0]. 
\end{align*}
If we think of the memory as a set $\mathbf{M}$ sitting on an $n$-dimensional Euclidean space $\mathbb{R}^n$, we can then think of these operations as $n \times n$ matrices.
From an algebraic point of view, we can realize these actions on the memory as left-monoid actions on a set, since matrix multiplication is associative and the matrix corresponding to the operation $\texttt{NO-OP}$ serves the role of an identity element in our computations. Below is the formulation of the Baby-NTM architecture:
\begin{align*}
\hiddenp{(t-1)} &= \hidden{(t-1)} + \weight{m} \mathbf{M}_{(t-1)}^{(0)} \\
\hidden{t} &= \text{tanh}(\weight{ih} x_t + \bias{ih} + \weight{hh} \hiddenp{(t-1)} + \bias{hh})\\
y_t &= \sigma (\weight{y} \hidden{t})\\
a_t &= \text{softmax} (\weight{a} \hidden{t}) \\
n_t &= \sigma (\weight{n} \hidden{t}) \\
\mathbf{M}_{t} &= \sum_{i=1}^{N} a_t ^{(i)} \cdot \left[\mathbf{OP}^{(i)}\right] \mathbf{M}_{t-1}\\
\mathbf{M}_{t}^{(0)} &= \mathbf{M}_{t}^{(0)} + n_t
\end{align*} 
\noindent where $\mathbf{M}_{t}$ denotes the memory configuration at time step $t$, $n_t$ the value of the element to be inserted to the first entry of the memory at time step $t$, $\mathbf{OP}^{(i)}$ the matrix corresponding to the $i$-th action on the memory
and $a_t ^{(i)}$ the weight of that action at time $t$, and all $\mathbf{W}$'s learnable matrices of the model.\footnote{As before, the memory here can be considered as the linear superposition of the results of all the memory operations.}

\subsection{Softmax Functions}
The softmax function in the calculation of $a_t$ in all these models enables us to map the values of the vector $\weight{a} \hidden{t}$ to a categorical probability distribution. We investigate the effect of more deterministic decisions about the stack/memory operations on the robustness of our model. A natural approach is to employ a softmax function with varying temperature $\tau$: 
\vspace{-0.7em}
\begin{align*}
    \text{softmax-temp}(x_{i}, \tau) = \frac{\exp(x_i/\tau)}{\sum_{j=1}^{N} \exp(x_j/\tau)}
\label{eqn:softmax-temp}
\end{align*}
The $\text{softmax-temp}$ function behaves exactly like the standard $\text{softmax}$ function when the temperature value $\tau$ equals $1$. As the temperature increases, $\text{softmax-temp}$ produces more uniform categorical class probabilities, whereas as the temperature decreases, the function outputs more discrete probabilities, like a one-hot encoding.

\begin{table*}[t]
\centering
\begin{tabular}{l | cccc | cccccccc}
\toprule
 \bf Sample & \multicolumn{4}{c|}{$( \ [ \ ] \ )$} & \multicolumn{7}{c}{$abc\#zyx$} \\ 
 \midrule
 \bf  Input &  $($&$[$&$]$&$)$& & $ a$&$b$&$c$&$\#$&$z$&$y$&$x$ \\ 
 \bf Output & $(/[/)$&$(/[/]$&$(/[/)$&$(/[$& & 
 $a/b/c/\#$&$a/b/c/\#$&$a/b/c/\#$&$z$&$y$&$x$&$\dashv$ \\ 
\bottomrule
\end{tabular}
\caption{Example input-output pairs for \dyck{2} (left) and the deterministic homomorphic palindrome language (right) under the sequence prediction paradigm.}
\label{tab:examples}
\vspace{-10pt}
\end{table*}

Furthermore, \citet{jang2016categorical} proposed an efficient and differentiable approximation to sampling from a discrete categorical distribution using a reparameterization trick:  
\vspace{-0.5em}
\begin{align*}
    \text{Gumbel-softmax-temp}(x_{i}, \{g_1, \ldots, g_{N}\}, \tau) \\= \frac{\exp((\log x_i + g_i)/\tau)}{\sum_{j=1}^{N} \exp((\log x_j + g_j)/\tau)}
\end{align*}

\noindent where $g_i \overset{\text{i.i.d.}}{\sim} \text{Gumbel} (0, 1)$.
As an alternative to the softmax function with varying temperature, one might want to use the Gumbel-softmax sampling method. In cases where we have more than two operations on the stack/memory, it might be tempting to prefer the Gumbel-softmax sampling approach for the calculation of $a_t$ values. We experiment with these alternatives below.

\vspace{-0.5em}
\section{Experimental Setup}
\vspace{-0.4em}
To evaluate the performance of the MARNNs, we conducted experiments on six computational tasks whose solutions require the formation of stack structures. In all the experiments, we used both standard and memory-augmented RNNs to explore the differences in their performances, in addition to the Stack-RNN model by \citet {joulin2015inferring}, and repeated each experiment $10$ times. Furthermore, we aimed to investigate softmax functions with varying temperature in our MARNNs and thus employed $12$ models with different configurations---two vanilla recurrent models, three MARNNs with three different softmax functions, and one Stack-RNN by \citet{joulin2015inferring}---for the six tasks.

 \subsection{The Sequence Prediction Task}
Following \citet{gers2001lstm}, we trained the networks as follows: At each time step, we presented one input symbol to the network and then asked the model to predict the set of next possible symbols in the language, based on the current symbol, the prior hidden states, and the stack. We used a one-hot representation to encode the inputs and a $k$-hot representation to encode the outputs. Table \ref{tab:examples} provides example input-output pairs for two of the experiments.

In all the experiments, the objective was to minimize the mean-squared error of the sequence predictions. We used an output threshold criterion of $0.5$ for the sigmoid layer ($y_t = \sigma (\cdot)$) to indicate which symbols were predicted by the model. Finally, we turned this sequence prediction task into a sequence classification task by \textit{accepting} a sequence if the model predicted \emph{all} of its output values correctly and \textit{rejecting} it otherwise. 

\vspace{-0.3em}
 \subsection{Training Details}
 \vspace{-0.3em}
In contrast to the models in \citet{joulin2015inferring, grefenstette2015learning, hao2018context, yu-etal-2019-learning}, our architectures are economical: Unless otherwise stated, the models are all single-layer networks with $8$ hidden units. In all the experiments, the entries of the memory were set to be one-dimensional, while the size of the memory in the Baby-NTMs was fixed to $104$ (since the length of the longest sequence in all the tasks was $100$). We used the Adam optimizer \citep{kingma2014adam}  and trained our models for three epochs. 

\vspace{-0.5em}
\section{Learning the Dyck Languages}
\vspace{-0.2em}
As described in the introduction, the Dyck languages \gendyck\ provide an ideal test-bed for exploring the ability of recurrent neural networks to capture the core properties of the context-free languages, their hierarchical modeling ability. None of the previous studies were able to learn the Dyck languages, with the exception of \dyck{1} (which can be captured using a simple one-counter machine). The main motivation of this paper was to introduce new neural architectures that could recognize \dyck{2} and other difficult context-free languages. 

\begin{table*}[t]
\small 
\centering
\begin{tabular}{c | c  c  c  c | c  c  c  c}
\toprule
& \multicolumn{4}{c|}{ \bf{Training Set}} & \multicolumn{4}{c}{ \bf{Test Set} } \\
\bf{Models}& \bf{Min} & \bf{Max} & \bf{Med} & \bf{Mean} & \bf{Min} & \bf{Max} & \bf{Med} & \bf{Mean} \\ \midrule
Vanilla RNN & 3.32 & 12.78 & 6.41 & 7.11 & 0 & 0 & 0 & 0 \\
Vanilla LSTM & 36.16 & 62.80 & 53.24 & 52.38 & 0.28 & 4.10 & 1.02 & 1.39 \\
Stack-RNN by \citetalias{joulin2015inferring} & 0 & 100 & 100 & 70.50 & 0 & 100 & 100 & 70.00 \\ 
\midrule
\bf{Stack-RNN+\textit{Softmax}} & \bf{100} & \bf{100} & \bf{100} & \bf{100} & \bf{99.96} & \bf{100} & \bf{100} & \bf{99.99} \\
\bf{Stack-RNN+\textit{Softmax-Temp}} & \bf{100} & \bf{100} & \bf{100} & \bf{100} & \bf{99.92} & \bf{100} & \bf{100} & \bf{99.99} \\
Stack-RNN+\it{Gumbel-Softmax} & 3.44 & 100 & 99.98 & 90.32 & 0 & 100 & 99.96 & 89.96 \\
\midrule
Stack-LSTM+\it{Softmax} & 62.52 & 100 & 100 & 95.69 & 2.78 & 100 & 98.25 & 87.51 \\
Stack-LSTM+\it{Softmax-Temp} & 46.70 & 100 & 100 & 94.67 & 0.80 & 100 & 99.73 & 89.84 \\
Stack-LSTM+\it{Gumbel-Softmax} & 50.26 & 100 & 99.94 & 94.97 & 0.70 & 99.94 & 99.33 & 88.68 \\
\midrule
Baby-NTM+\it{Softmax} & 2.56 & 100 & 100 & 75.80 & 0 & 100 & 99.91 & 68.73 \\
Baby-NTM+\it{Softmax-Temp} & 1.16 & 100 & 99.88 & 72.43 & 0 & 100 & 96.97 & 68.23 \\
Baby-NTM+\it{Gumbel-Softmax} & 5.66 & 100 & 99.88 & 89.39 & 0 & 99.90 & 99.54 & 86.85 \\
\bottomrule
\end{tabular}
\caption{The performances of the vanilla and memory-augmented recurrent models on \dyck{2}. Min/Max/Median/Mean results were obtained from $10$ different runs of each model with the same random seed across each run. We note that both Stack-RNN+\textit{Softmax} and Stack-RNN+\textit{Softmax-Temp} achieved full accuracy on the test sets in $8$ out of $10$ times.}
\label{tab:results_dyck}
\end{table*}

\subsection{The \dyck{2} Language} 
We trained the Stack-RNN, Stack-LSTM, and Baby-NTM architectures with slightly different memory-controller configurations, in addition to standard RNNs, to learn \dyck{2}. 

A probabilistic context-free grammar for \dyck{2} can be written as follows:
\begin{align*}
S \rightarrow \begin{cases} 
(\, S\, ) & \text{with probability } \frac{p}{2} \\
[\, S\, ] & \text{with probability } \frac{p}{2} \\
S\,S & \text{with probability } q \\ 
\varepsilon & \text{with probability } 1 - (p+q) 
\end{cases}
\end{align*}
\noindent where $0 < p, q < 1$ and $p+q < 1$.

\begin{figure}[t]
\centering
{\includegraphics[width=0.50\textwidth]{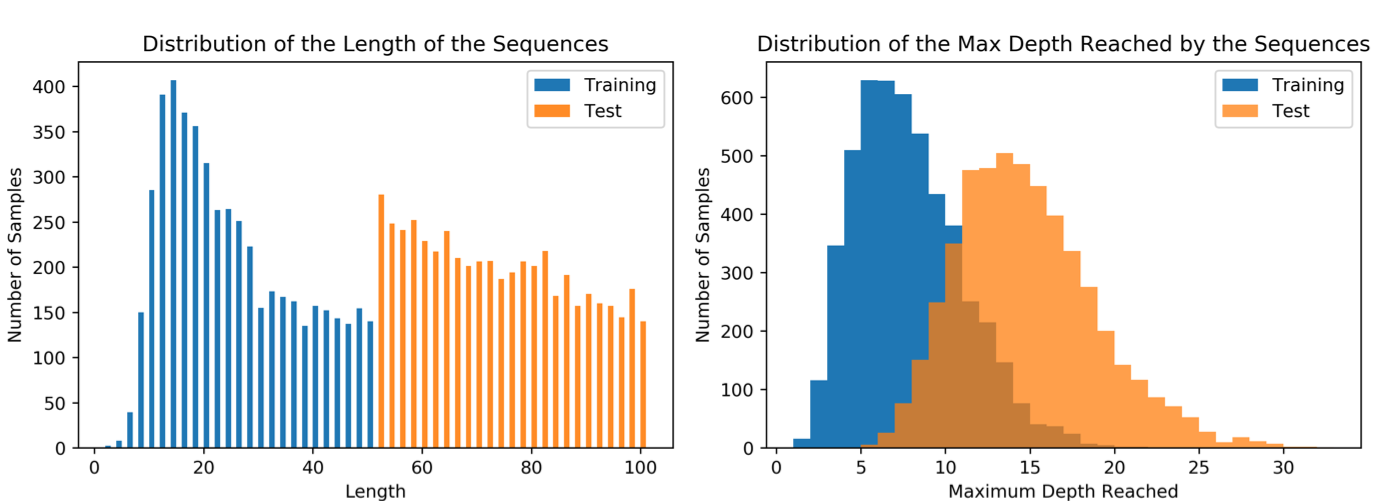}}
\caption{Length and maximum depth distributions of training/test sets for an example \dyck{2} 
experiment.
}
\label{fig:dyck2}
\end{figure}

Setting $p = \frac{1}{2}$ and $q = \frac{1}{4}$, we generated $5000$ distinct Dyck words, whose lengths were bounded to $[2, 50]$, for the training sets. Similarly, we generated $5000$ distinct words whose lengths were bounded to $[52, 100]$ for the test sets. Hence, there was no overlap between the training and test sets. Test set performance requires generalization well past the training set lengths. As it can be seen in the length and maximum depth distributions of the training and test sets for one of the \dyck{2} experiments in Figure \ref{fig:dyck2}, the test samples contained longer dependencies than the training sample.

\begin{figure*}[t!]
\centering
{\includegraphics[width=.98\textwidth]{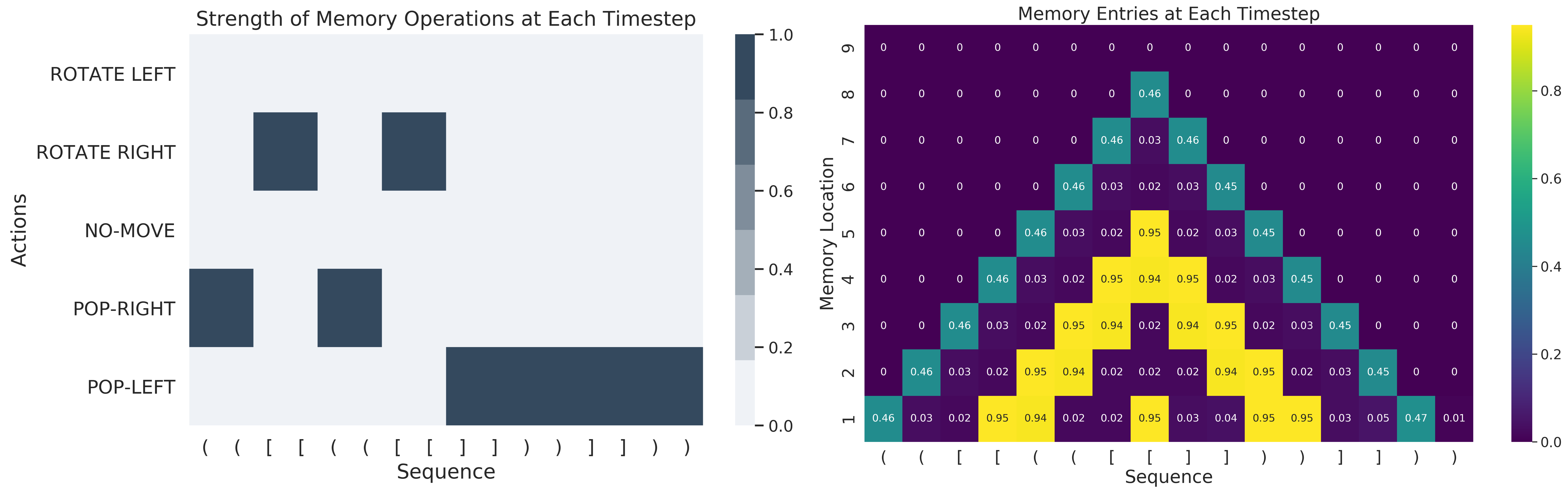}}
\caption{Visualizations of the strength of the memory operations (left) and the values of the memory entries (right) of a Baby-NTM+\textit{Softmax} model trained to learn \dyck{2}. We highlight that the Baby-NTM appears to have learned to emulate a simple but effective differentiable pushdown automaton to recognize \dyck{2}.}
\label{fig:ntm_viz}
\end{figure*}

Table~\ref{tab:results_dyck} lists the performances of the vanilla and memory-augmented recurrent models on the training and test sets for the Dyck language. Our empirical results highlight the dramatic performance difference between the memory-augmented recurrent networks and vanilla recurrent networks: We note that almost all our stack/memory-augmented architectures achieved full accuracy on the test set, which contained longer and deeper sequences than the training set, while the vanilla RNNs and LSTMs failed to generalize with below $5\%$ accuracy. We further observe that the Stack-RNN proposed by \citet{joulin2015inferring} performed nearly as well as our models, though ours performed better than theirs on average.

When evaluated based on their empirical median and mean percent-wise performances, the Stack-RNNs appear to be slightly more successful than the Stack-LSTMs and the Baby-NTMs. Both the Stack-RNN+\textit{Softmax} and Stack-RNN+\textit{Softmax-Temp} obtained perfect accuracy on the test sets $8$ out of $10$ times, whereas the best Stack-LSTM variant, Stack-LSTM+\textit{Softmax-Temp}, achieved perfect accuracy only $3$ out of $10$ times. Nevertheless, we acknowledge that most of our stack-augmented models were able to successfully generalize well beyond the training data. \footnote{We additionally note that, contrary to our initial expectation, using a softmax activation function with varying temperature did not improve the performance of our memory-augmented neural models in general. However, the networks might actually benefit from temperature-based softmax functions in the presence of more categorical choices, because currently the models have only a very limited number of memory operations.}

\begin{table*}[h!]
\small 
\centering
\begin{tabular}{c | c  c  c  c | c  c  c  c}
\toprule
& \multicolumn{4}{c|}{ \bf{Training Set}} & \multicolumn{4}{c}{ \bf{Test Set} } \\ 
\bf{Models} & \bf{Min} & \bf{Max} & \bf{Med} & \bf{Mean} & \bf{Min} & \bf{Max} & \bf{Med} & \bf{Mean} \\ 
\midrule
Vanilla RNN & 0.82 & 14.88 & 11.19 & 9.52 & 0 & 0 & 0 & 0 \\
Vanilla LSTM & 24.16 & 39.76& 31.55 & 32.58 & 0 & 0.16 & 0.02 & 0.04 \\
Stack-RNN by \citetalias{joulin2015inferring} & 9.02 & 100 & 98.17 & 79.32 & 0 & 100 & 91.29 & 66.72 \\ 
\midrule
\bf{Stack-RNN+\textit{Softmax}} & \bf{7.80} & \bf{100} & \bf{100} & \bf{81.75} & \bf{0} & \bf{100} & \bf{100} & \bf{80.00} \\
Stack-RNN+\it{Softmax-Temp} & 37.64 & 99.98 & 95.74 & 81.95 & 0.06 & 98.18 & 67.32 & 52.49 \\
Stack-RNN+\it{Gumbel-Softmax} & 1.78 & 100 & 44.55 & 50.71 & 0 & 99.98 & 21.94 & 43.65 \\
\midrule
Stack-LSTM+\it{Softmax} & 33.98 & 100 & 92.25 & 77.97 & 0.04 & 99.94 & 61.54 & 55.49 \\
Stack-LSTM+\it{Softmax-Temp} & 37.64 & 99.98 & 95.74 & 81.95 & 0.06 & 98.18 & 67.32 & 52.49 \\
Stack-LSTM+\it{Gumbel-Softmax} & 25.74 & 99.98 & 78.21 & 72.01 & 0 & 99.2 & 27.08 & 42.17 \\
\midrule
Baby-NTM+\it{Softmax} & 4.60 & 100 & 84.29 & 60.63 & 0 & 100 & 23.44 & 44.51 \\
Baby-NTM+\it{Softmax-Temp} & 6.40 & 100 & 16.44 & 39.97 & 0 & 100 & 0.51 & 27.46 \\
Baby-NTM+\it{Gumbel-Softmax} & 0.76 & 100 & 11.70 & 43.42 & 0 & 99.9 & 0 & 38.76 \\
\bottomrule
\end{tabular}
\caption{The performances of the vanilla and memory-augmented recurrent models on \dyck{3}. In $32$ out of $100$ trials, the MARNNs with $8$ hidden units and one-dimensional memory achieved over $99\%$ accuracy on the test sets. However, increasing the dimensional of the memory for our MARNNs further improved our results.}
\label{tab:results_dyck3}
\end{table*}

\begin{figure*}[h!]
\centering
{\includegraphics[width=.98\textwidth]{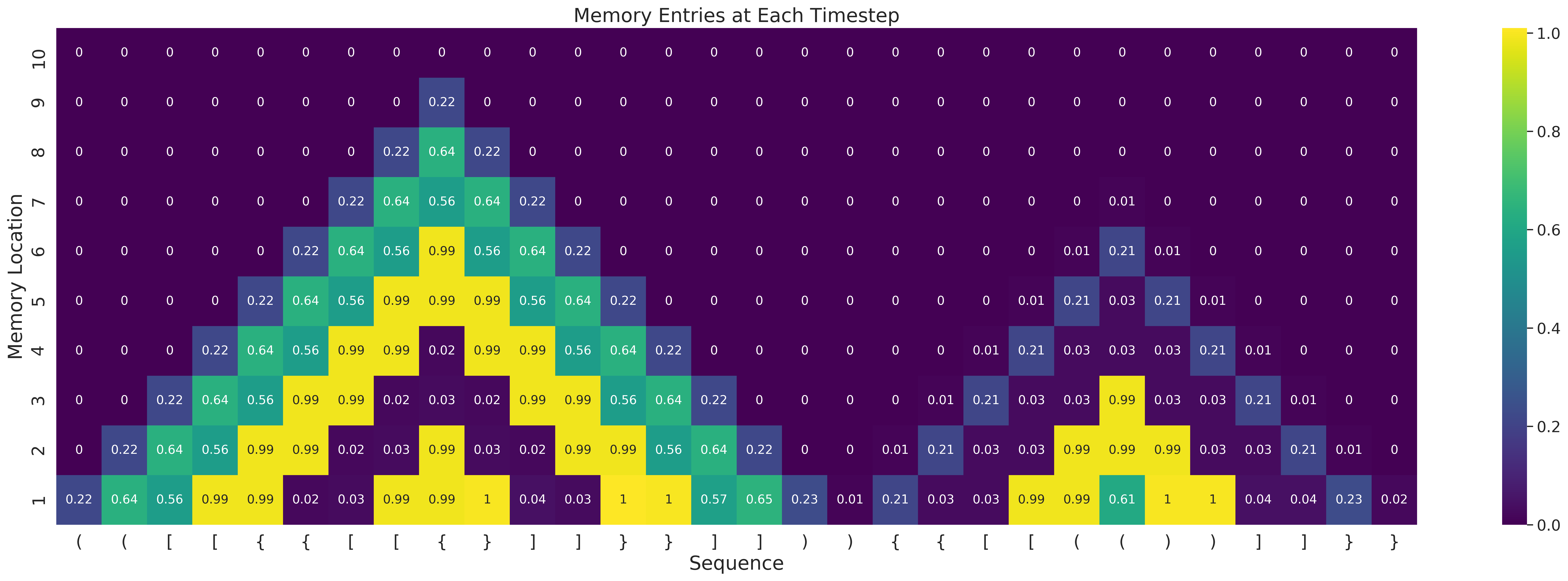}}
\vspace{-0.2em}
\caption{Visualization of the values of the memory entries of a Baby-NTM+\textit{Softmax} model trained to learn \dyck{3}.}
\label{fig:ntm_viz_dyck3}
\end{figure*}

Figure~\ref{fig:ntm_viz} provides a visualization of the strengths of the memory operations and the change in the values of the entries of the memory component of one of our memory-augmented models (a Baby-NTM+\textit{Softmax} with $8$ hidden units) at each time step when the model was presented a sample in \dyck{2}.
The Baby-NTM appears to be using its \texttt{ROTATE-RIGHT} and \texttt{POP-RIGHT} operations for the open parentheses `$($' and `$[$', respectively, and \texttt{POP-LEFT} operation for both of the closing parentheses `$)$' and `$]$', thereby emulating a simple PDA-like mechanism. A careful inspection of the memory entries of the Baby-NTM indicates that the model utilizes a special marker with a distinct value (${\sim}0.45$ in our example) to distinguish an empty stack configuration from a processed stack configuration. On the other hand, the memory alone does not dictate the output values: The hidden states of the model govern the overall behavior and embody a finite-state control, as shown in the formulation of the Baby-NTM.\footnote{The visualizations for the other memory-augmented models were qualitatively similar, though some networks learned more complex representations. We further emphasize that the dimensions of the external stack and memory entries in our MARNNs were set up to be one-dimensional for visualization purposes, but we additionally experimented with higher dimensional memory structures and observed that such additions often increased the overall performances of the models, especially the performance of the Baby-NTMs.} 

\begin{table*}[t!]
\small 
\centering
\begin{tabular}{c | c  c  c  c | c  c  c  c}
\toprule
& \multicolumn{4}{c|}{ \bf{Training Set}} & \multicolumn{4}{c}{ \bf{Test Set} } \\ 
\bf{Models} & \bf{Min} & \bf{Max} & \bf{Med} & \bf{Mean} & \bf{Min} & \bf{Max} & \bf{Med} & \bf{Mean} \\ \midrule
Vanilla RNN & 21.19 & 25.71 & 23.53 & 23.39 & 0 & 0.02 & 0 & 0 \\
Vanilla LSTM & 32.47 & 41.62 & 37.35 & 37.05 & 0 & 0.06 & 0 & 0.01 \\
Stack-RNN by \citetalias{joulin2015inferring} & 99.47 & 100 & 100 & 99.94 & 97.60 & 100 & 99.99 & 99.70 \\
\midrule
\bf{Stack-RNN+\textit{Softmax}} & \bf{99.92} & \bf{100} & \bf{100} & \bf{99.99} & \bf{99.32} & \bf{100} & \bf{99.99} & \bf{99.85} \\
Stack-RNN+\it{Softmax-Temp} & 36.83 & 100 & 98.54 & 80.09 & 0 & 100 & 78.44 & 60.88 \\
Stack-RNN+\it{Gumbel-Softmax} & 20.90 & 99.98 & 99.93 & 91.62 & 0 & 99.92 & 99.50 & 87.69 \\
\midrule
Stack-LSTM+\it{Softmax} & 98.48 & 100 & 99.99 & 99.79 & 91.12 & 100 & 99.18 & 98.23 \\
Stack-LSTM+\it{Softmax-Temp} & 36.83 & 100 & 98.54 & 80.09 & 0 & 100 & 78.44 & 60.88 \\
Stack-LSTM+\it{Gumbel-Softmax} & 36.20 & 99.94 & 67.50 & 68.62 & 0 & 99.90 & 24.61 & 44.47 \\
\midrule
\bf{Baby-NTM+\textit{Softmax}} & \bf{99.94} & \bf{100} & \bf{100} & \bf{99.99} & \bf{99.00} & \bf{100} & \bf{99.97} & \bf{99.87} \\
Baby-NTM+\it{Softmax-Temp} & 86.12 & 100 & 100 & 98.15 & 8.56 & 100 & 99.91 & 88.40 \\
Baby-NTM+\it{Gumbel-Softmax} & 22.56 & 99.98 & 99.86 & 75.46 & 0 & 99.86 & 99.23 & 63.49 \\
\bottomrule
\end{tabular}
\caption{The performances of the vanilla and memory-augmented recurrent models  on the \dyck{6}. We note that the MARNNs in this example contain $12$ hidden units and $5$-dimensional external stack/memory. In $70$ out of $100$ trials, the MARNNs performed over $99\%$ accuracy. Overall, the Baby-NTM+\textit{Softmax} had the best performance.}
\label{tab:dyck6}
\end{table*}

\subsection{The \dyck{3} and \dyck{6} Languages}
We further conducted experiments on the \dyck{3} and \dyck{6} languages to evaluate the ability of our memory-augmented architectures to encode more complex hierarchical representations. The training and test corpora were generated in the same style as the previous task; however, we included $15,000$ samples in the training set for \dyck{6}, due to its complexity. 

As shown in Table~\ref{tab:results_dyck3}, the Stack-RNN+\textit{Softmax} model had the best performance among all the neural networks on the \dyck{3} learning task, obtaining perfect accuracy in eight out of ten trials. Following the Stack-RNNs, the Stack-LSTMs and Baby-NTMs, on average, achieved around $50\%$ and $37\%$ accuracy on the test set, respectively. On the other hand, the Stack-RNN by \citet{joulin2015inferring} could generalize better than most of our models in terms of its median and mean scores, albeit still not better than our Stack-RNN.

Figure~\ref{fig:ntm_viz_dyck3} illustrates the behavior of one of our Baby-NTMs as the model is presented a long sequence in \dyck{3}. It is remarkable to witness how the memory-augmented model makes use of its external memory to learn a sequence of actions to recognize a sample in \dyck{3}. Similar to the behavior of the previous model in Figure~\ref{fig:ntm_viz}, the RNN controller of the Baby-NTM model in this instance appears to be using the differentiable memory as a stack-like structure and inserting distinct values to the memory at different time steps. Furthermore, we note the presence of special markers ($0.22$ in the first half and $0.21$ in the second half -- both colored blue) in the memory: These idiosyncratic memory elements marking the bottom of the used portion of the stack enable the model to know when to predict \emph{only} the set of open parentheses. 

In contrast, the overall performance of the MARNNs for \dyck{6} was much lower than for \dyck{2} and \dyck{3}. For instance, none of our models could obtain full accuracy on the training or test sets; the maximum score our models could achieve was $60.38\%$. We wondered whether increasing the dimension of the memory would remedy the problem. Table~\ref{tab:dyck6} summarizes our new results with the same architectures containing $12$ hidden units and $5$-dimensional augmented stack/memory. We saw a significant increase in the performance of our models: In $60$ out of $90$ trials, our enhanced MARNNs achieved almost perfect ($\geq 99\%$) accuracy on the test set.

\vspace{-0.6em}
\section{Learning Palindrome Languages}
\vspace{-0.3em}
Our previous results established that the MARNNs can learn Dyck languages, which represent the core of the CFL class. We note that the Dyck languages incorporate a notion of palindrome: The intersection of \dyck{n} with $p^* \bar{p}^*$ leads to a definition of a specific type of a homomorphic palindrome language $w\varphi(w^R)$, where $^*$ is the Kleene star, $\varphi$ a homomorphism given by $p_i \mapsto \bar{p}_i$, and $w^R$ the reversal of $w$. Therefore, we would expect our models to be able to learn various deterministic versions of palindrome languages.

\begin{table*}[t!]
\small 
\centering
\begin{tabular}{c | c  c  c  c | c  c  c  c}
\toprule
& \multicolumn{4}{c|}{ \bf{Training Set}} & \multicolumn{4}{c}{ \bf{Test Set} } \\ 
\bf{Models} & \bf{Min} & \bf{Max} & \bf{Med} & \bf{Mean} & \bf{Min} & \bf{Max} & \bf{Med} & \bf{Mean} \\ \midrule
Vanilla RNN & 0 & 0  & 0 & 0 & 0 & 0 & 0 & 0 \\
Vanilla LSTM & 0 & 5.22  & 2.23 & 2.39 & 0 & 0  & 0 & 0 \\
Stack-RNN by \citetalias{joulin2015inferring}& 0 & 100 & 46.04 & 49.13 & 0 & 100 & 50.42 & 50.00 \\
\midrule
Stack-RNN+\it{Softmax} & 0 & 100 & 99.99 & 60.00 & 0 & 100 & 100 & 60.00 \\
Stack-RNN+\it{Softmax-Temp} & 0 & 100 & 100 & 70.00 & 0 & 100 & 100 & 70.00 \\
Stack-RNN+\it{Gumbel-Softmax} & 0 & 100 & 17.10 & 43.42 & 0 & 100 & 16.98 & 43.39 \\
\midrule
Stack-LSTM+\it{Softmax} & 0 & 100 & 100 & 61.36 & 0 & 100 & 100 & 60.00 \\
Stack-LSTM+\it{Softmax-Temp} & 0 & 100 & 100 & 70.07 & 0 & 100 & 100 & 70.00 \\
\bf{Stack-LSTM+\textit{Gumbel-Softmax}} & \bf{0} & \bf{100} & \bf{100} & \bf{80.20} & \bf{0} & \bf{100} & \bf{99.98} & \bf{79.99} \\
\midrule
Baby-NTM+\it{Softmax} & 0 & 100 & 67.16 & 53.43 & 0 & 100 & 66.55 & 53.31 \\
Baby-NTM+\it{Softmax-Temp} & 0 & 100 & 99.99 & 60.00 & 0 & 100 & 100 & 60.00 \\
Baby-NTM+\it{Gumbel-Softmax} & 0 & 100 & 60.43 & 52.09 & 0 & 100 & 61.30 & 52.26 \\
\bottomrule
\end{tabular}
\caption{The performances of the vanilla and memory-augmented recurrent models on the deterministic homomorphic palindrome language. Most of our MARNNs achieved almost full accuracy on the test sets.}
\label{tab:results_hom_palindrome}
\end{table*}

\vspace{-0.4em}
\subsection{Homomorphic Palindrome Language} 
Our first target of exploration is the deterministic homomorphic palindrome language, the language of words $w\#\varphi(w^{R})$ where $w \in \{a, b, c\}^*$, $\#$ is a symbol serving to mark the center of the palindrome, and $\varphi$ maps $a$ to $x$, $b$ to $y$, and $c$ to $z$. We use the notion of recognition from the previous section, predicting at each symbol the set of all possible following symbols. Viewed as a transduction, this amounts to the following task:
\vspace{-0.2em}
\begin{align*}
    w\# \varphi(w^R) &\Rightarrow (a/b/c/\#)^{|w|} \varphi(w^{R})\dashv
\end{align*}
\vspace{-0.2em}
The training set for this task contained $5000$ unique samples of length varying from $2$ to $50$, and the test set contained $5000$ unique samples of length varying from $52$ to $100$. We remark that there was no overlap between the training and test sets, just as in the case of the Dyck language tasks. 

Table~\ref{tab:results_hom_palindrome} lists the performances of the vanilla and memory-augmented models on the deterministic homomorphic palindrome language. We highlight the success of our models once again: While our MARNNs often performed with perfect accuracy on the training and test sets, the standard recurrent models could not predict even one sample in the test set correctly. Overall, most of the variants of the Stack-RNN/LSTM and Baby-NTM models seem to have learned how to emulate pushdown automata: They learned to push certain values into their memory or stack whenever they read a character from the $\{a, b, c\}$ alphabet and then started popping them one by one after they would see $\#$, and at the last step, the model predicted the end of the sequence token $\dashv$. The models did not perform equally well though: When evaluated on their mean and median percent-wise performances, for instance, the Stack-LSTMs were found to generalize better than the Stack-RNNs and the Baby-NTMs in this task. Further, it is hard to make a conclusive statement about whether employing a softmax function with varying temperature in our MARNNs had any benefit. Nevertheless, the Stack-LSTMs+\textit{Gumbel-Softmax} performed slightly better than the other models in terms of their mean percentages on the test sets.
 
\begin{table*}[t!]
\small 
\centering
\begin{tabular}{c | c  c  c  c | c  c  c  c}
\toprule
& \multicolumn{4}{c|}{ \bf{Training Set}} & \multicolumn{4}{c}{ \bf{Test Set} } \\ 
\bf{Models} & \bf{Min} & \bf{Max} & \bf{Med} & \bf{Mean} & \bf{Min} & \bf{Max} & \bf{Med} & \bf{Mean} \\ \midrule
Vanilla RNN & 0.06 & 0.90 & 0.46 & 0.46 & 0 & 0 & 0 & 0 \\
Vanilla LSTM & 0.68 & 5.08 & 3.62 & 3.50 & 0 & 0 & 0 & 0 \\
Stack-RNN by \citetalias{joulin2015inferring} & 0.16 & 100 & 50.29 & 50.19 & 0 & 100 & 50.00 & 50.00 \\
\midrule
Stack-RNN+\it{Softmax} & 0.38 & 100 & 100 & 77.39 & 0 & 100 & 100 & 76.65 \\
Stack-RNN+\it{Softmax-Temp} & 0.14 & 100 & 100 & 80.05 & 0 & 100 & 100 & 80.00 \\
Stack-RNN+\it{Gumbel-Softmax} & 0.18 & 100 & 99.98 & 77.71 & 0 & 100 & 99.98 & 77.83 \\
\midrule
Stack-LSTM+\it{Softmax} & 2.02 & 100 & 100 & 80.60 & 0 & 100 & 100 & 79.99 \\
Stack-LSTM+\it{Softmax-Temp} & 0.06 & 100 & 100 & 70.49 & 0 & 100 & 100 & 70.00 \\
Stack-LSTM+\it{Gumbel-Softmax} & 2.18 & 100 & 100 & 80.48 & 0 & 100 & 100 & 80.00 \\
\midrule
Baby-NTM+\it{Softmax} & 0.08 & 100 & 100 & 86.65 & 0 & 100 & 100 & 86.67 \\
Baby-NTM+\it{Softmax-Temp} & 0 & 100 & 100 & 70.07 & 0 & 100 & 100 & 70.00 \\
\bf{Baby-NTM+\textit{Gumbel-Softmax}} & \bf{0.20} & \bf{100} & \bf{100} & \bf{90.01} & \bf{0} & \bf{100} & \bf{99.97} & \bf{89.96} \\
\bottomrule
\end{tabular}
\caption{The performances of the vanilla and memory-augmented recurrent models on the string reversal task under the transduction setting. In $32$ out of $90$ trials, our MARNNs obtained perfect accuracy.}\label{tab:results_string_reversal}
\end{table*}
 
 \vspace{-0.6em} 
\subsection{Simple Palindrome Language}
\vspace{-0.2em}

Taking the homomorphism $\varphi$ to be the identity map in the previous language, it is reasonable to expect the models to learn the $w\# w^R$ palindrome language. We evaluated recognition of this language once again as a possible-next-symbol prediction task, which can be viewed as the following sequence transduction task:
\vspace{-0.4em}
\begin{align*}
    w\# w^R &\Rightarrow (a/b/c/\#)^{|w|} w^{R}\dashv
\end{align*}
Surprisingly, all of our MARNN models had difficulty learning this language. Only in three of $90$ trials were our MARNNs able to learn the language; other times, the models typically obtained $0\%$ accuracy during testing. When we increased the dimensionality of the memory to five, however, our MARNNs immediately learned the task with almost full accuracy again. Given that the only difference between the previous task and this task is the second half of the strings, we conjectured that our models were getting confused in the second half: Because of vocabulary overlap in the two halves, the models might be using information from the second half when predicting the set of possible symbols in the second half, thereby getting distracted and finding themselves stuck at bad local minima. To verify our hypothesis, we thus performed one more task, string reversal, which we describe in the following section.

\vspace{-0.5em}
\section{Learning the String-Reversal Task}
In the previous section, we witnessed a strange phenomenon: Our MARNN models with $8$ hidden units and one-dimensional external memory could learn the deterministic homomorphic palindrome language, but not the simple palindrome language. Since the only difference between the two tasks is the existence of a non-trivial isomorphism $\varphi$ (and the vocabulary overlap in the two halves), we wanted to perform the string reversal task under a sequence transduction setting in which the reversal appears only in the output:
\begin{align*}
   w\#^{|w|} &\mapsto \#^{|w|} w^{R}
\end{align*}
The training and test sets were similar to the previous cases: $5000$ samples each, with lengths bounded by $[2,50]$ and $[52, 100]$, respectively.

Table~\ref{tab:results_string_reversal} illustrates that most of our MARNNs achieved perfect accuracy on the test sets in this version of the string reversal task. The results corroborated our conjecture by showing that when the second half of the input samples contained symbols from an alphabet other than the one used in the first half (in this case, the $\#$ symbol), the memory-augmented models do not get confused and act in the desired way (pushing elements into the stack in the first half and popping them one by one in the second half after seeing the marker $\#$). When we visualized the hidden states and the memory entries of the models for this task, we observed that our MARNNs learned to emulate simple pushdown-automata.

\section{Conclusion}
In this paper, we introduced three memory-augmented neural architectures and provided the first demonstration of neural networks learning to recognize the generalized Dyck languages, which represent the ``core'' of the context-free language class. We further evaluated the learning capabilities of our models on recognizing the deterministic homomorphic palindrome language and simple palindrome language under the sequence prediction framework and performing the string-reversal task under a sequence transduction setting. In all the experiments, our MARNNs outperformed the vanilla RNN and LSTM models and often attained perfect accuracy on both the training and test sets. 

Since we limited the dimensionality of the external memory in our memory-augmented architectures to one, we were also able to visualize the changes in the external memory of the Baby-NTMs trained to learn the \dyck{2} and \dyck{3} languages. Our simple analysis revealed that our MARNNs learned to emulate pushdown-automata to recognize these Dyck languages. \citet{hao2018context} mention that their Neural-Stack models could not perfectly employ stack-based strategies to learn an appropriate representation to recognize the \dyck{2} language, and further address the difficulty of training stack-augmented recurrent networks. Although we agree that it is challenging to train MARNNs due to various optimization issues, one can still train these models with as few as eight or twelve hidden units to learn the Dyck languages, and our empirical findings support this claim. 

\section{Acknowledgment}
The authors appreciate the helpful comments of Michael Hahn, Yoav Goldberg, Drew Pendergrass, Dan Stefan Eniceicu, and Filippos Sytilidis. M.S.\ gratefully acknowledges the support of the Harvard College Research Program (HCRP) and the Harvard Center for Research on Computation and Society Research Fellowship for Undergraduate Students. S.G. was supported by a Siebel Fellowship. Y.B.\ was supported by the Harvard Mind, Brain, and Behavior Initiative. The computations in this paper were run on the Odyssey cluster supported by the FAS Division of Science, Research Computing Group at Harvard University.

\bibliography{paper}
\bibliographystyle{acl_natbib}

\newpage

\onecolumn

\appendix
\section{Comparison of Stack-RNN architectures}
\label{sec:appendix}

Recall the formulation of our Stack-RNN architecture in Section~\ref{sec:stack-rnn}. We update $\hidden{t}$, the hidden state at time $t$, as follows: 
\begin{align*}
    \hidden{t} &= \text{tanh}(\weight{ih} x_t + \bias{ih} + \weight{hh} \hiddenp{(t-1)} + \bias{hh})
\end{align*}
\noindent where $\hiddenp{(t-1)}$ is defined to be:
\begin{align*}
    \hiddenp{(t-1)} &= \hidden{(t-1)} + \weight{sh} s_{(t-1)}^{(0)}
\end{align*}
Rewriting the equation for $\hidden{t}$, we realize that our formulation of Stack-RNN is almost equivalent to the Stack-RNN model by \citet{joulin2015inferring}:
\begin{align*}
    \hidden{t} &= \text{tanh}(\weight{ih} x_t + \bias{ih} + \weight{hh} \hiddenp{(t-1)} + \bias{hh})\\
    &= \text{tanh}(\weight{ih} x_t + \bias{ih} + \weight{hh} (\hidden{(t-1)} + \weight{sh} s_{(t-1)}^{(0)}) + \bias{hh})\\
    &= \text{tanh}(\weight{ih} x_t + \bias{ih} + \weight{hh} \hidden{(t-1)} + \underbrace{\weight{hh} \weight{sh}}_{(*)} s_{(t-1)}^{(0)} + \bias{hh})
\end{align*}
In our Stack-RNN architecture, $s_{(t-1)}^{(0)}$ depends on $(*)$, namely $\weight{hh} \weight{sh}$, whereas in \citeauthor{joulin2015inferring}'s Stack-RNN model, it only depends on $\weight{sh}$. Furthermore, we make use of $\tanh (\cdot)$, instead of $\sigma (\cdot)$, to achieve non-linearity and include bias terms, namely $\bias{ih}$ and $\bias{hh}$, in our definition of $\hidden{t}$.
\end{document}
